\documentclass[10pt,twocolumn,a4paper]{article}
\usepackage[utf8]{inputenc}
\usepackage{url}
\usepackage{graphicx}
\usepackage{authblk}

\usepackage{algorithm}
\usepackage{algorithmic}
\usepackage[font={footnotesize}]{caption}

\usepackage{amsmath,amssymb,amsfonts,latexsym,amscd,epsf,psfrag} 
\def\bbbr{{\mathbb R}} 

\usepackage{hyperref}

\renewcommand\footnotemark{}

\title{\vspace{-4ex}The problems with using STNs to align CNN feature maps}

\author{Lukas Finnveden}
\author{Ylva Jansson\thanks{\hspace{-4.5ex}The support from the Swedish Research Council (contract 2018-03586) is gratefully acknowledged. \\Corresponding author: yjansson@kth.se.}}
\author{Tony Lindeberg}
\affil{KTH Royal Institute of Technology, Stockholm, Sweden}
\date{\vspace{-5ex}}

\begin{document}
\maketitle

\begin{abstract}  
Spatial transformer networks (STNs) were designed to enable CNNs to learn invariance to image transformations. 
STNs were originally proposed to transform \emph{CNN feature maps} as well as input images. 
This enables the use of more complex features when predicting transformation parameters.
However, since STNs perform \emph{a purely spatial transformation}, they do not, in the general case, have the ability to align the feature maps of a transformed image and its original. We present a theoretical argument for this and investigate the practical implications, showing that this inability is coupled with decreased classification accuracy. We advocate taking advantage of more complex features in deeper layers by instead sharing parameters  between the classification and the localisation network. 
\end{abstract}
\vspace{-1ex}


\section{Theory}
\paragraph{Spatial transformer networks} (STNs) \cite{JadSimZisKav-NIPS2015,LinLuc-CVPR2017} were introduced as an option for CNNs to learn invariance to image transformations by transforming input images or convolutional feature maps before further processing. A spatial transformer (ST) module is composed of a localization network that predicts transformation parameters and a transformer that transforms an image or \emph{a feature map} using these parameters. An STN is a network with one or several ST modules 
at arbitrary depths. 

An ST module can clearly be used for \emph{pose alignment} of images when applied directly to the input. Assume an input image $f:\bbbr^n \mapsto \bbbr$ and a set of image transformations $T_g$ indexed by some parameter $g$. Transformed images $T_g f$ could be transformed into a canonical pose if the ST module correctly learns to apply the inverse transformation: $ T_g^{-1} T_g f = f$.

\begin{figure}[htbp]
\begin{center}
\includegraphics[width=0.4\textwidth]{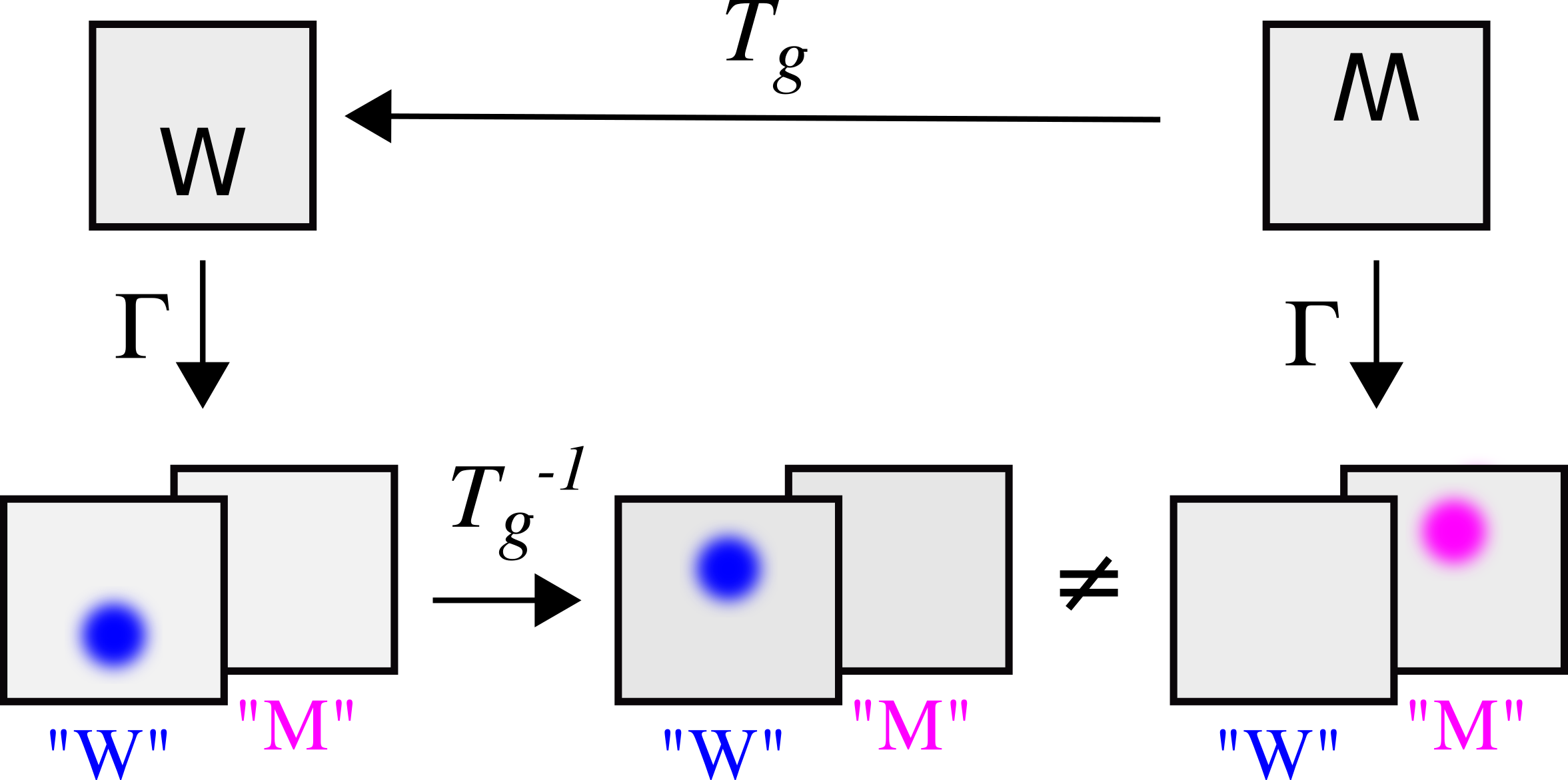}
\end{center}
\caption{Inversely transforming \emph{the feature map} will, in general, not align the feature maps of a transformed image and those of its original. The network $\Gamma$ has two feature channels "W" and "M". $T_g$ corresponds to a 180$^\circ$ rotation.}
\label{fig:tiny-proof}
\end{figure}

However, if applying the inverse spatial transformation to \emph{a convolutional feature map} $(\Gamma f)(x,c)$, here with $c$ channels, this will, in the general case, not result in alignment of the feature maps of a transformed image and those of the original image
\begin{equation}
    T_g^{-1} (\Gamma\, T_g f)(x,c) \neq (\Gamma f)(x,c)
\end{equation}
The intuition for this is illustrated in Figure~\ref{fig:tiny-proof}, where $\Gamma$ has two feature channels for recognising the letters "W" and "M". Note how a purely spatial transformation cannot align the feature maps $\Gamma f$ and $\Gamma\, T_g f$, since there is also a shift in \emph{the channel dimension}.
A similar reasoning applies to a wide range of spatial image transformations.

This gives rise to the question of the relative benefits of transforming the input vs. transforming intermediate feature maps in STNs. Is there a point in transforming intermediate feature maps if it cannot support invariant recognition?


\vspace{-3ex}



\section{Experiments}
To investigate the practical implications of the inability of ST modules to support invariance, if applied to CNN feature maps, we compared 4 different network configurations on rotated and translated MNIST and the Street View House Numbers dataset (SVHN): \textbf{(i)} A standard CNN \textit{(CNN)} \textbf{(ii)} An STN with the ST module directly following the input \textit{(STN-C0)} \textbf{(iii)} An STN with the ST module following convolutional layer X \textit{(STN-CX)} and \textbf{(iv)} An STN which transforms the input but where the localization network \emph{shares the first X layers} with the classification network, which enables the use of more complex features to infer the transformation parameters \textit{(STN-SLX}). 

\begin{figure}
	\centering
	\includegraphics[height=0.25\textwidth]{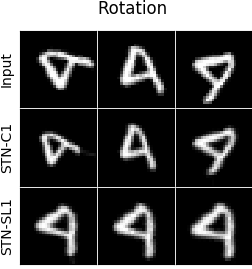}
	\includegraphics[height=0.25\textwidth]{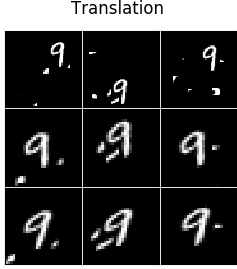}
	\caption{Visualisation of image/feature map alignment for rotated and translated MNIST images (top rows). \mbox{STN-C1} fails to compensate for rotations but performs better for translations (middle rows). STN-SL1 finds a canonical pose both for rotated and translated images (bottom rows).}
	\label{fig:rotate-vs-translate}
\end{figure}

\begin{figure}
	\centering
	\includegraphics[height=0.25\textwidth]{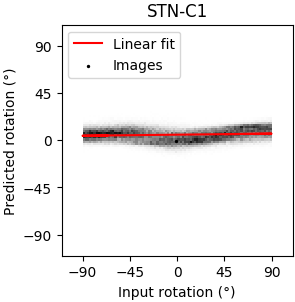} 
	\includegraphics[height=0.25\textwidth]{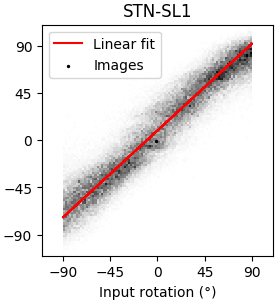} \\
	\caption{The rotation angle \emph{predicted} by the ST module for MNIST images as a function of the rotation \emph{applied} to the input image. STN-C1 has not learned to predict the image orientation (left). The reason for this is that a rotation is, in fact, not enough to align deeper layer feature maps. This is because a rotation of the feature map does not correspond to a rotation of the input. STN-SL1, which transforms the input, correctly predicts the image orientation (right).}
	\label{fig:angle-predictions}
\end{figure}

Figure \ref{fig:rotate-vs-translate} and Figure \ref{fig:angle-predictions} demonstrate that the transformation learned by STN-C1 does
not correspond to pose alignment of rotated input images, while the transformation learned by STN-SL1 does. 
For translations, STN-C1 performs better, since a translation 
does not imply a shift in the feature map channel dimension. Thus STN-C1 works better as \emph{an attention mechanism} than to compensate for image transformations.
Table~\ref{tab:mnist} shows that the inability of STN-C1 to align feature maps of rotated images leads to decreased classification performance. Table 2 shows that, while STN-CX suffers from a tradeoff between using deeper layer features and its inability to support invariance, STN-SLX can fully take advantage of deeper features.


\vspace{-2ex}

\section{Conclusions}
We have investigated the practical implications of the inability of an STN to align \emph{CNN feature maps} to enable invariant recognition. Our results show that this inability is clearly visible in practice and, indeed, negatively impacts classification performance. When more complex features are needed to correctly estimate an image transformation, we thus advocate using deeper layer features by means of parameter sharing but, importantly, still \emph{transform the input}. 
Our results also has implications for other similar approaches that are designed to compensate for image transformations with spatial transformations of \emph{CNN feature maps or filters}.



\begin{table}[h]
\caption{Classification error on rotated and translated MNIST data for the different network versions.}
\begin{tabular}{rcc}
\hline
Network & Rotation & Translation \\
\hline
CNN & $1.71\%$  & $1.72\% $ \\
STN-C0 & $1.08\% $ & $\textbf{1.08\%} $ \\
STN-C1 & $1.32\% $ & $1.15\% $ \\
STN-SL1 & $\textbf{0.98\%} $ & $1.10\%$ \\ 
\end{tabular}
\label{tab:mnist}
\end{table}
\vspace{-1ex}
\begin{table}[h]
\caption{Classification error on the SVHN dataset 
when transforming intermediate feature maps at different depths vs transforming the input but using parameter sharing between the localisation and the classification network.}
\begin{tabular}{rcc}
\hline
Depth & STN-CX & STN-SLX  \\
\hline
X=0 & 3.81\% & 3.81\% \\
X=3 & 3.70\% & 3.54\%  \\ 
X=6 & 3.91\% & 3.29\% \\ 
X=8 & 4.00\% & \textbf{3.27\%} \\ 
\end{tabular}
\label{tab:svhn}
\end{table}

\bibliographystyle{abbrv}
\footnotesize
\vspace{-3ex}
\bibliography{ext_abst}

\end{document}